\documentclass[10pt,twocolumn,letterpaper]{article}

\usepackage{cvpr}
\usepackage{times}
\usepackage{epsfig}
\usepackage{graphicx}
\usepackage{amsmath}
\usepackage{amssymb}

\newcommand{\argmax}{\operatornamewithlimits{argmax}}

\usepackage[breaklinks=true,bookmarks=false]{hyperref}

\cvprfinalcopy 

\def\cvprPaperID{895} 

\begin{document}

\title{Recognizing Car Fluents from Video}


\author{Bo Li$^{1,*}$, Tianfu Wu$^{2}$, Caiming Xiong$^{3,}$\thanks{This work was done when Bo Li was a visiting student and Caiming Xiong was a Postdoc at UCLA.} \ and Song-Chun Zhu$^{2}$ \\
$^{1}$Beijing Lab of Intelligent Information Technology, Beijing Institute of Technology\\
$^{2}$Department of Statistics, University of California, Los Angeles  \ \ \ \ \ \ \ \ \ \ \ \   $^{3}$Metamind Inc.\\  
\small{\it boli86@bit.edu.cn, \{tfwu, sczhu\}@stat.ucla.edu, cmxiong@metamind.io}}
\maketitle

\begin{abstract}
\vspace{-2mm}
Physical fluents, a term originally used by Newton~\cite{newton}, refers to time-varying object states in dynamic scenes. 
In this paper, we are interested in inferring the fluents of vehicles from video. For example, a door (hood, trunk) is open or closed through various actions, light is blinking to turn. 
Recognizing these fluents has broad applications, yet have received scant attention in the computer vision literature. 
Car fluent recognition entails a unified framework for car detection, car part localization and part status recognition, which is made difficult by large structural and appearance variations, low resolutions and occlusions. This paper learns a spatial-temporal And-Or hierarchical model to represent car fluents. The learning of this model is formulated under the latent structural SVM framework. Since there are no publicly related dataset, we collect and annotate a car fluent dataset consisting of car videos with diverse fluents. In experiments, the proposed method outperforms several highly related baseline methods in terms of car fluent recognition and car part localization. 
\vspace{-3mm}
\end{abstract}

\vspace{-2mm}
\section{Introduction} \label{sec:intro}

\vspace{-1mm}
\subsection{Motivation and Objective}
\begin{figure}
\centering
{\includegraphics[width = 0.48\textwidth]{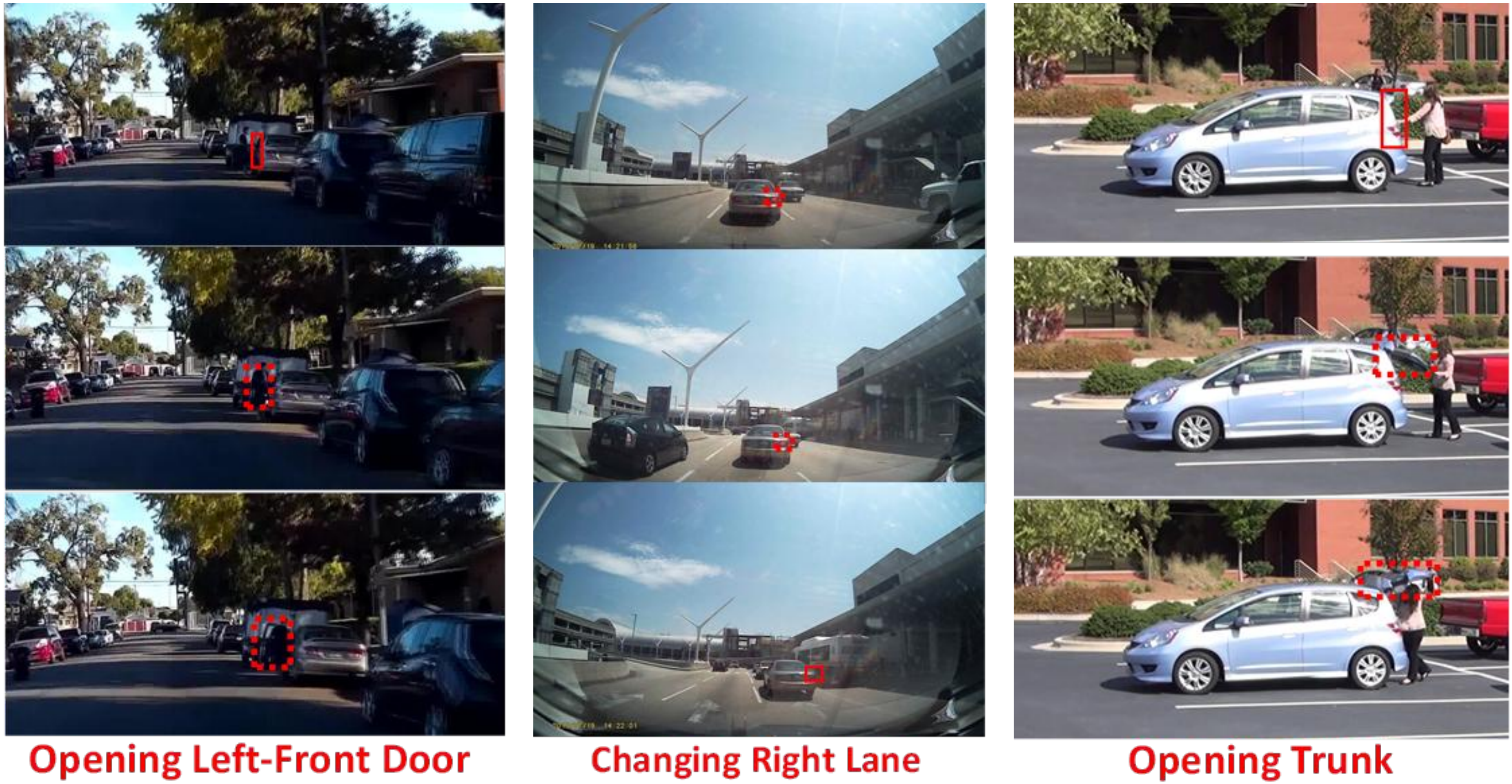}}
\caption{Some important applications of car fluent recognition. Best viewed in color and zoom in. \vspace{-5mm} }
\label{fig:app}
\end{figure}

\vspace{-1mm}
The term of physical fluents is first introduced by Newton \cite{newton} to represent the time-varying statuses of object states. In the commonsense literature, it is defined as the specifically varied object status in a time sequence \cite{mueller}. 
As a typical instance, car fluent recognition has applications in video surveillance and self-driving.
In autonomous driving, car fluents are very important to infer the road conditions and the intents of other drivers. The left of Fig. \ref{fig:app} shows an example, when seeing the left-front door of a roadside car is opening, the autonomous car should slow down, pass with cautions, or be ready to stop.
In the freeway context, the autonomous car should give way when the front car is blinking to require a left or right lane change (see the middle figure in Fig.~\ref{fig:app}).
In the parking lot scenario, car part status change indicates info for reasoning about human-car interactions and surveillance, e.g., a woman is opening car trunk to pick up stuff (see the right of Fig. \ref{fig:app}).  
In general, fluents recognition is also essential in inferring the minds of human and robotics in Cognition, AI and Robotics.

\begin{figure}
\centering
{\includegraphics[width = 0.48\textwidth]{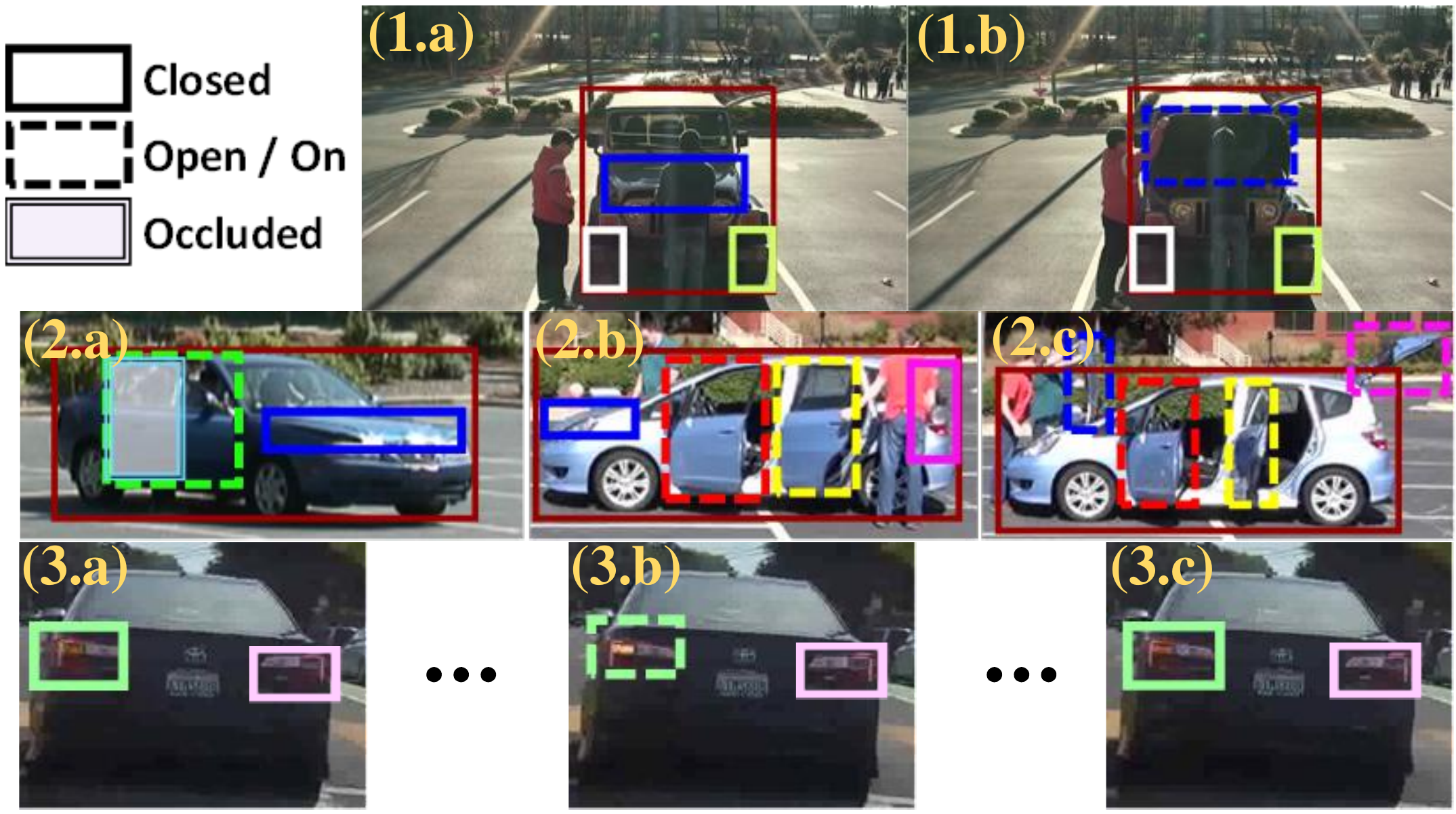}}
\caption{Illustration of the large structural and appearance variations and heavy occlusions in car fluent recognition. We focus on the fluents of functional car parts (e.g., doors, trunks, lights) in this paper. For clarity, we show the cropped single cars only. See Fig. \ref{fig:app} and Fig. \ref{fig:snap} for the whole context. See text for details.\vspace{-5mm} }
\label{fig:part_status_variation}
\end{figure}

While there is a vast literature in computer vision on vehicle detection and viewpoint estimation \cite{DPM,pff_grammar,bojan_cvpr12,bojan_cvpr13,xiang_cvpr15,OhnBar,kunhe,its_c,its_e}, 
reasoning about the time-varying states of objects (i.e.,  fluents) has been rarely studied.  
Car fluents recognition is a hard and unexplored problem. As illustrated in Fig.~\ref{fig:part_status_variation}, car fluents have large structural and appearance variations, low resolutions, and severe occlusions, which present difficulties at multiple levels. In this paper, we address the following tasks in a hierarchical spatial-temporal And-Or Graph (ST-AOG).

\textit{i) Detecting cars} with different part statuses and occlusions (e.g., in Fig.\ref{fig:part_status_variation} (1.b), the frontal-view jeep with hood being open and persons wandering in the front. In the popular car datasets (such as the PASCAL VOC car dataset~\cite{pascal} and KITTI car dataset~\cite{kitti}), most cars do not have open parts (e.g., open trunk/hood/door) together with car-person occlusions. So, a more expressive model is needed in detecting cars in car fluent recognition. 

\textit{ii) Localizing car parts} which have low detectability as individual parts (e.g., the open hood and trunk in Fig.~\ref{fig:part_status_variation} (2.c), the tail lights in the last row of Fig.~\ref{fig:part_status_variation}) or even are occluded (e.g., the trunk in Fig.~\ref{fig:part_status_variation} (2.b)). This situation is similar, in spirit, to the well-known challenge of localizing hands/feet in human pose recovery~\cite{ramananPose}. Temporal contextual information is needed to localize those parts, besides a spatial hierarchical car pose model. 
 
\textit{iii) Recognizing time-varying car part statuses} is a new task. Unlike object attributes (such as gender and race) which are stable for a long time, the time-varying nature of car fluents presents more ambiguities (e.g., in Fig.~\ref{fig:part_status_variation} (3.a-3.c), the periodical status change of car lights, and its ambiguous appearance and geometric features). Unlike action recognition which focuses on humans and does not account for the pose and statuses of parts~\cite{laptev08,dense_traj}, car fluent recognition is fine-grained in both spatial and temporal domains. 

\begin{figure*}
\centering
{\includegraphics[width = 0.90\textwidth]{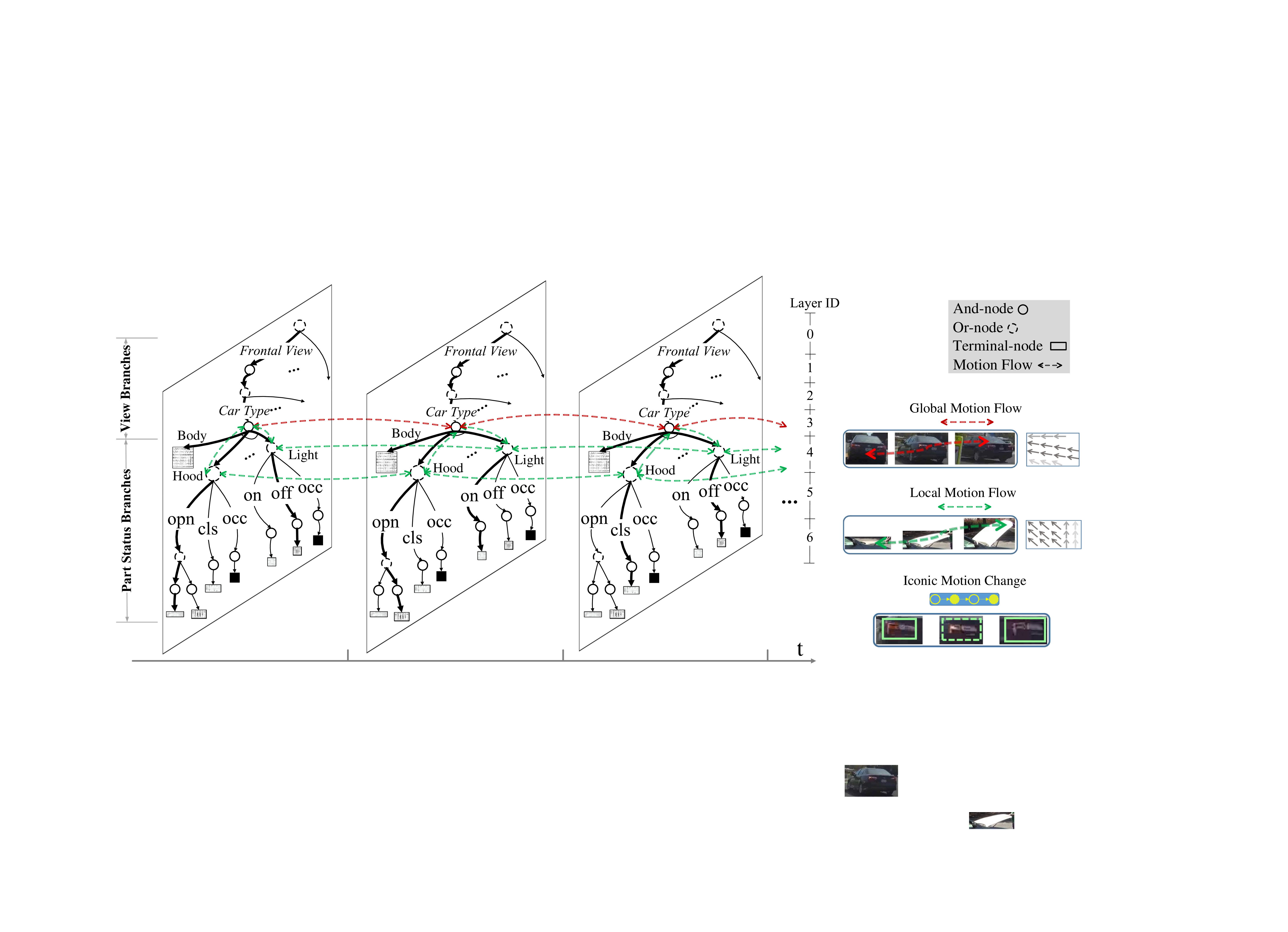}}
\caption{Illustration of the proposed ST-AOG for representing car fluents. In space (vertical), the ST-AOG represents cars by quantized views, types, and semantic parts with different statuses. This is used to model appearance and geometry info. In time (horizontal), the ST-AOG represents temporal transitions by $3$ type of motion flows: Global motion flow, local motion flow and iconic motion change, which are illustrated on the right. Here, ``opn", ``cls", ``occ", ``on", ``off" denotes ``open", ``close", ``occluded", ``turn on", ``turn off", respectively. For each fluent video, our ST-AOG outputs a parse graph consists of frame-level parse trees, which are shown by the bold arrows.
We omit the temporal transitions of Terminal-nodes for clarity. \vspace{-5mm}}
\label{fig:ST-AOG}
\end{figure*}

In this paper, we propose to learn a ST-AOG \cite{zhu_grammar,pff_grammar,leozhu_nips07,gupta09} to represent car fluents at semantic part-level. Fig. \ref{fig:ST-AOG} shows a portion of our model, the ST-AOG span in both spatial and temporal dimensions. In space, it represents the whole car, semantic parts, part status from top to bottom. In time, it represents the location and status transitions of the whole car and car parts from left to right.
Given a test video, ST-AOG will output frame-level car bounding boxes, semantic part (e.g, door, light) bounding boxes, part statuses (e.g., open/close, turn on/off), and video-level car fluents (e.g., opening trunk, turning left).

Because of the lateral edges in Fig. \ref{fig:ST-AOG}, our proposed ST-AOG is no longer a directed acyclic graph (DAG), thus dynamic programming (DP) cannot be directly utilized. To cope with this problem, we incorporate loopy belief propagation (LBP) \cite{Weiss00} and DP in our inference, and adopt a part-based hidden Markov Model (HMM)  in temporal transition for each semantic part (see Fig.~\ref{fig:lbp_hmm}). 
All the appearance, deformation and motion parameters in our model are trained jointly under the latent structural SVM framework.

\begin{figure*}
\centering
\includegraphics[width = 1.0\textwidth]{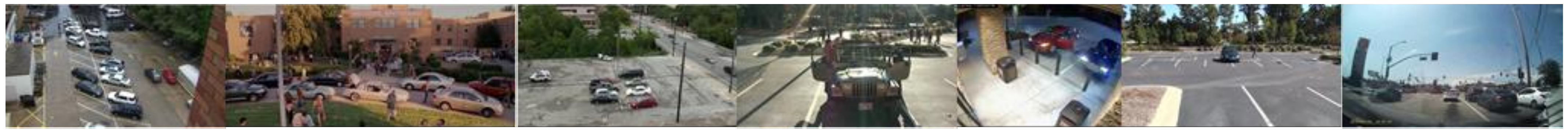}
\caption{Sample images in Car-Fluent dataset. \vspace{-5mm} }
\label{fig:snap} 
\vspace{-3mm}
\end{figure*}

To train and evaluate the ST-AOG, we collect and annotate a car fluent dataset due to the lack of publicly available benchmark (Fig.~\ref{fig:snap} shows some examples, and details of the dataset will be given in Section \ref{sec:dataset}). 
Experimental results show that our model is capable of jointly recognizing semantic car fluents, part positions, and part status in video. It outperforms several highly related state-of-the-art baselines in fluent recognition and part localization (sliding window-based). Besides, our model can be directly incorporated with popular improved dense trajectories (iDT) \cite{idt} and C3D features \cite{c3d} to encodes more general information.

\vspace{-2mm}
\section{Related Work and Our Contributions}
\vspace{-1.5mm}
We briefly review the most related topics in $3$ streams:

\textbf{Car Detection and View Estimation.} In computer vision and intelligent transportation, there is a considerable body of work on car detection and view estimation \cite{DPM,pff_grammar,bojan_cvpr12,bojan_cvpr13,xiang_cvpr15,OhnBar,junli,kunhe,its_c,its_e}. 
Though those works have successfully improve the performance on popular benchmarks \cite{pascal,kitti}, the output is usually the car bounding box and a quantized rough view. Some other works \cite{michael_stark,zia,xiaoou_car15} aim to get car configurations with detailed or fine-grained output to describe the more meaningful car shape, rather than a bounding box. 
However, all of those works generally regarded car as a static rigid object, while pay little attention to the functionalities of semantic car parts and the fact that cars presenting large geometry and appearance transformations during car fluents change. 

\textbf{Video-based Tasks in Vision, Cognition, AI and Robotics.} In computer vision, a significant effort has been devoted to video-based tasks: event \cite{event_feifei,event_nevatia,event_xu}, action detection/recognition \cite{survey_action,Zuffi,jiangwang2,xiaohan} and pose estimation \cite{ramananPose,Andriluka,Ferrari09}. These papers are related to our work, however, most of them are based on human models, while very little work has done on car fluents and related car part status estimation. 
In cognitive science, the concept of fluents is mainly used to represent the object status changing in time series \cite{mueller}, and it has been used in causality inference \cite{mueller,Heckerman,Brand, amy}. Furthermore, fluents are also related to action-planning in AI and Robotics \cite{blum,Yang_LFA_15}.

\textbf{Human-Object Recognition.} As mentioned in \cite{amy}, the status of object fluent change is caused by some other objects. In our case, the car fluent change is mainly conducted by humans. Thus our work is related to part recognition \cite{dickinson,dickinson2} and human-object interactions \cite{Ryoo1,Ryoo2,Gupta,bangpeng,poselets09,visual_phrases,desai_eccv12}. But, here we jointly and explicitly model car detection, part localization and status estimation. 

\textit{Our Contributions.} This paper makes three contributions to car fluent recognition: 
 
i) It presents a ST-AOG to represent the spatial-temporal context in car fluent recognition. Car fluent recognition is a new task in the computer vision literature.  
 
ii) It presents a new car fluent dataset with per-frame annotations of cars, car types, semantic parts, part statuses and viewpoints.
	
iii) It provides a new method for car fluent recognition and part status estimation, and outperforms several state-of-the-art baseline methods.

\vspace{-2mm}
\section{Representation}
\vspace{-1mm}
\subsection{Spatial-Temporal And-Or Graph}
\vspace{-1mm}
Fig.~\ref{fig:ST-AOG} illustrates our ST-AOG, it models a car fluent by the viewpoint, car type, semantic parts and corresponding part statuses hierarchically.
In addition to the appearance and deformation, it also models the temporal transitions of the semantic parts.

Utilizing $I_i$ and $I_{i+1}$ to denote neighbouring frames, the ST-AOG is defined by a $3$-tuple, 
\vspace{-2mm}
\begin{equation}
\small
\mathcal{G} = (V, E, \Theta) 
\vspace{-3mm}
\end{equation}
where $V$ is the node set consisting of a set of non-terminal nodes (i.e., And-nodes $V_{And}$ and Or-nodes $V_{Or}$) and a set $V_T$ of terminal-nodes (i.e., appearance templates). 
$E = (E_{\mathcal{S}},E_{\mathcal{M}})$ is the edge set maintaining the structure of the graph. Specifically, $E_{\mathcal{S}}$ is the spacial edges accounting for composition and deformation, $E_{\mathcal{M}}$ is the lateral edges accounting for temporal transitions. $\Theta = (\Theta^{app}, \Theta^{def}, \Theta_{\mathcal{M}}, \Theta^{bias})$ is the set of parameters, and we have,   

\textit{i) the appearance parameters $\Theta^{app}=(\Theta^{app}_t, t\in V_T)$} consists of the appearance templates for all terminal-nodes. Our model is general, in which any type of features can be used. In this paper, we use the histogram of oriented gradient (HOG) feature~\cite{HOG} and convolutional neural network (CNN) feature \cite{lecun89} as the appearance feature for illustration. 

\textit{ii) the deformation parameters $\Theta^{def}$} are defined for all in-edges of all terminal-nodes, $(v, t)\in E$ ($v\in V_{And} \cup V_{Or}$ and $t\in V_T$), to capture the local deformation when placing a terminal-node $t$ in a feature pyramid. For the deformation feature, we adopt the standard separable quadratic function of the displacement of $t$ with respect to its parent node $v$, $(dx, dx^2, dy, dy^2)$, as done in the deformable part-based models~\cite{DPM,ssdpm} and the car AOG model~\cite{boli_eccv14}.   

\textit{iii) the transition parameters $\Theta_{\mathcal{M}}$} are defined on the single car and semantic part nodes. They are used to weight the effects of temporal transition between $I_i$ and $I_{i+1}$. 
As illustrated on the right of Fig. \ref{fig:ST-AOG}, we compute $3$ types of motion flows for car fluents: 

\textbf{a) global motion flow.} We compute the optical flow of the whole car (e.g., body), and utilize it as the global feature to localize car in time. This global feature is very helpful for moving cars, especially in the far view, in which the semantic part is too small to be detected.

\textbf{b) local motion flow.} We compute the optical flow of ``mobile" car part (e.g., door), and utilize it as the temporal feature to recognize the position and status of the part. Here, the ``mobile" car part is the part that can be open or closed, which has motion features during fluent change.

\textbf{c) iconic motion flow.} For car lights, the most distinctive feature is its intensity change and temporally periodical turn on/off statuses. This is related to the work in iconic change \cite{black98,Black2000} and change detection \cite{change_det}. Here, we use the appearance and optical flow templates to localize car lights, and compute the temporal intensities (vector of normalized intensities) to recognize the periodical light status change (e.g., left-turn/right turn).

For optical flow, we compute the pyramid feature to cope with different scales.

\textit{iv) the bias parameters $\Theta^{bias}$} are defined for the child nodes of all Or-nodes to encode either the prior (e.g., different appearance templates for the ``open" part), or the compatibility of parsing scores among different sub-categories (e.g., different viewpoints of a jeep).


Although the hierarchical structure of our ST-AOG is similar to those spatial-only models used in~\cite{boli_eccv14,bojan_cvpr13} for car detection, we introduce the semantic part Or-nodes (to be used to define fluent) for detailed part status modelling in both spatial and temporal domains.

A \textbf{parse tree}, $pt$, is an instantiation of the ST-AOG being placed at a location in the spatial-temporal feature pyramid. It is computed by following the breadth-first-search order of the nodes in $\mathcal{G}$, selecting the best child node at each encountered Or-node (based on the scoring function to be defined later on). 
The bold black edges in Fig. \ref{fig:ST-AOG} show the parse trees of ST-AOG on three neighbouring video frames.
For a video with $N$ frames, we can get its \textbf{parse graph} as $pg = \{pt^{i,i+1}\}_{i=1}^{N-1}$.

Based on the $pg$, we extract frame-level part bounding boxes and statues, and utilize them to design spatial-temporal features for fluent recognition. To capture long-term spatial-temporal info, our model can also integrate iDT \cite{idt} and C3D features \cite{c3d} (see Section \ref{sec:recog}).

\vspace{-1mm}
\subsection{The Scoring Function}
\vspace{-1mm}

The scoring function of ST-AOG is recursively defined w.r.t. its structure. For simplicity, we will use $v$ and $\widetilde{v}$ ($v \in V$) to represent the temporally neighbouring nodes (on frames $I_i$ and $I_{i+1}$) below.

Let $O \in V_{Or}$ denotes the Or-node in the ST-AOG, $A \in V_{And}$ be the parent node of a terminal-node $t\in V_T$. We model $t$ by a 4-tuple $(\theta_t^{app}, \theta_{t|A}^{def}, \sigma_t, a_{t|A})$  where $\sigma_t$ is the scale factor for placing $t$ w.r.t. $A$ in the feature pyramid ($\sigma_t\in\{0,1\}$), and $a_{t|A}$ is the anchor position of $t$ relative to $A$. 

i) Given $A, \widetilde{t}$ and their locations $l_A, l_{\widetilde{t}}$, \textit{the scoring function} of placing $t$ at the position $l_t$ is then defined by,   
\vspace{-2mm}
\small
\begin{align} 
& \nonumber S(t|l_{\widetilde{t}}, A, l_A) =  \max_{l_t}[<\theta_t^{app}, \Phi^{app}(l_t)> - \\
&  <\theta_{t|A}^{def}, \Phi^{def}(l_t, l_A)> + \theta^{T}_{\mathcal{M}} \lVert l_t - l_{\widetilde{t}} + \mathcal{F}(l_t) \rVert^{2}_{2}]
\vspace{-1mm}
\end{align}
\normalsize
where $\Phi^{app}(l_t)$ is the appearance features (HOG or CNN) and $\Phi^{def}(l_t, l_A)$ is the deformation features. $\theta^{T}_{\mathcal{M}}$ is the motion flow weight of $t$. $\mathcal{F}(l)$ is the motion flow between frames $I_i$ and $I_{i+1}$ computed at position $l$.

ii) Given $\widetilde{A}$ and its location $l_{\widetilde{A}}$ at frame $I_{i+1}$, \textit{the scoring function of $A$} is defined by, 
\vspace{-3mm}
\small
\begin{align}
& \nonumber S(A, l_A | \widetilde{A}, l_{\widetilde{A}}) = \sum_{v\in ch(A)} [S(v|A,l_A) + b_A \\
& \qquad \qquad + \theta_{\mathcal{M}}^{A} \lVert l_{v|A} - l_{v|\widetilde{A}} + \mathcal{F}(l_{v|A}) \rVert^{2}_{2}]
\vspace{-1mm}
\end{align}
\normalsize
where $ch(A)$ is the set of child nodes of $A$, $b_A$ is the bias parameter. $\theta_{\mathcal{M}}^{A}$ is the motion flow weight of $A$. 

iii) Given $\widetilde{O}$ and its location $l_{\widetilde{O}}$ at frame $I_{i+1}$, \textit{the scoring function of $O$} is defined by, 
\vspace{-2mm}
\small
\begin{align}
& \nonumber S(O, l_O | \widetilde{O}, l_{\widetilde{O}}) = \max_{v\in ch(O)} [S(v|O,l_O) + \\
& \qquad \qquad \theta_{\mathcal{M}}^{O} \lVert l_{v|O} - l_{v|\widetilde{O}} + \mathcal{F}(l_{v|O}) \rVert^{2}_{2}] \label{eqn:score_o}
\vspace{-1mm}
\end{align}
\normalsize
where $ch(O)$ is the set of child nodes of $O$, $l_O$ is the position of $O$. $\theta_{\mathcal{M}}^{O}$ is the optical flow weight of $O$. 

For \textbf{temporal term} $\lVert l_v - l_{\widetilde{v}} + \mathcal{F}(l_v) \rVert^{2}_{2}, v \in V$, since the scoring function of $v$ is conditioned on $\widetilde{v}$ in time, while $\widetilde{v}$ is conditioned on its child nodes or parent node in space, there are loops in this structure, and we couldn't use DP to compute above scoring functions directly. As similar to \cite{mixbody}, we resort to LBP \cite{Weiss00} to get an approximate solution.
For computational efficiency, the motion flows are computed only at nodes corresponds to the car and its semantic parts. For other nodes, their motion flow are implicitly embedded, as their score maps are related to nodes with temporal links, and their scoring functions are computed as above but without the temporal terms.

Given a fluent video clip with $N$ frames, the overall score of ST-AOG will be:
\vspace{-4mm}
\small
\begin{align}
& S(I_{1:N}|pg,\mathcal{G}) = \frac{1}{N-1} \sum_{i=1}^{N-1} S^{i,i+1}(O,l_O|\widetilde{O},l_{\widetilde{O}}) \label{eqn:poster}
\vspace{-3mm}
\end{align} 
\normalsize
where $O$ is the root Or-node in ST-AOG. In probabilistic model, Eqn. (\ref{eqn:poster}) can be interpreted as a log-posterior probability up to a constant. Our objective is to maximize Eqn. (\ref{eqn:poster}), given all appearance, deformation, and motion features from training videos. 


\vspace{-2mm}
\section{Inference by LBP and DP} \label{sec:infer}
\vspace{-1mm}
To cope with the loop introduced by motion transition, we extend the traditional inference procedure in AOG \cite{boli_eccv14,tangram} with LBP. Given an input video, Our inference procedure includes $4$ steps:

\textit{i)} For each frame $I_i$, we omit the temporal links between it and its neighbour frames, and compute the appearance feature pyramid, optical flow pyramid, and score maps for all nodes in Layer $3-6$ in the ST-AOG by the Depth-First-Search (DFS) order; This step is similar to the inference algorithm in \cite{boli_eccv14}.

\begin{figure}
\centering
{\includegraphics[width = 0.48\textwidth]{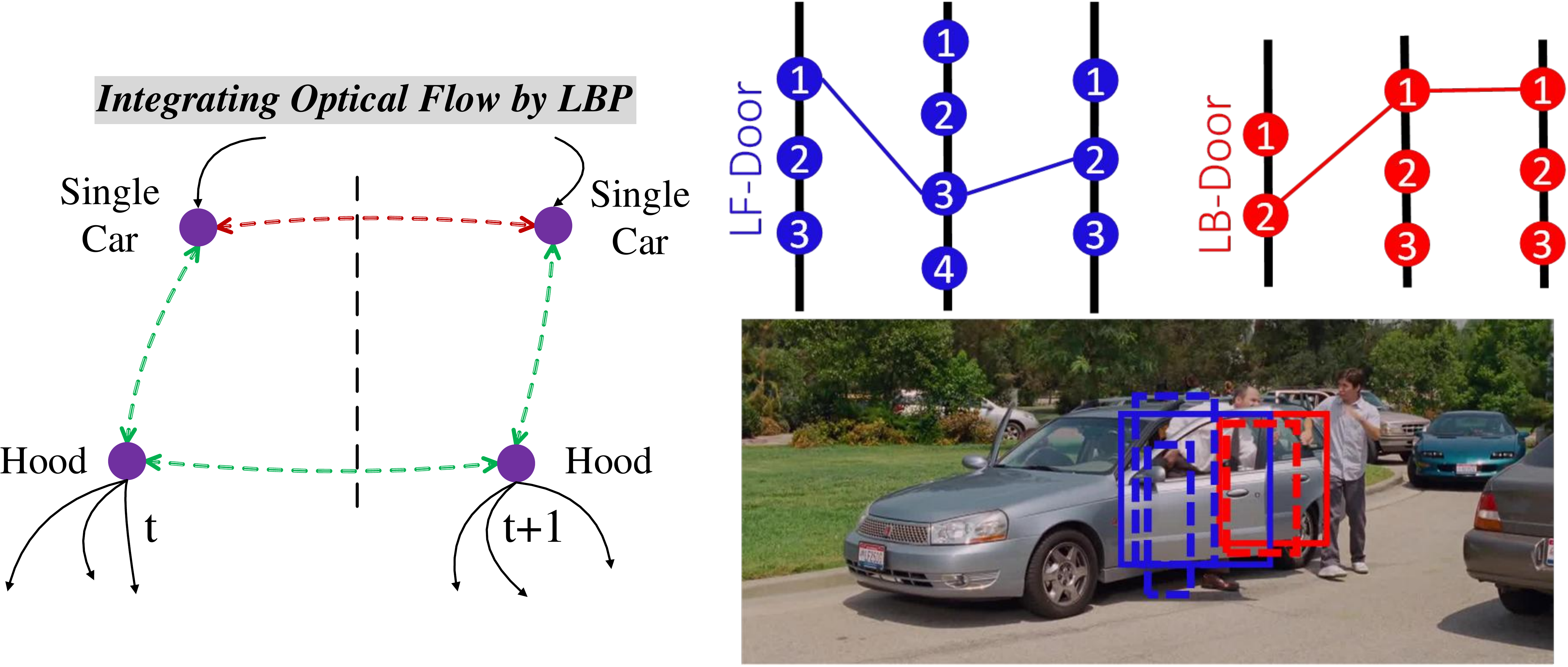}}
\caption{Left: Computing optical flow with LBP. For simplicity, we just show the flow circle of ``Body-Hood". Right: Semantic part inference with part-based HMM. 
Here, each bold line represents a frame, circles represent part proposals with different status, thin lines represent the temporal transitions, and semantic parts are represented by different colors.
For simplicity, we just show the bounding box proposals of the left-front door and left-back door. Best viewed in color and zoom in. }
\label{fig:lbp_hmm}
\vspace{-5mm}
\end{figure}

\textit{ii)} After we get the score maps of the semantic part Or-nodes and the single car And-nodes, we further integrate the score maps with optical flow maps, as can be seen on the left of Fig. \ref{fig:lbp_hmm}. For each semantic part $p$ with its parent root node $r$, we focus on four nodes, i.e., $p$, $r$, $\widetilde{p}$, $\widetilde{r}$, and omit other spatial links connected to them. At the beginning, we send message from $p$ to $r$, then from $r$ to the rest to update the message. When the difference of the propagated message in two consecutive iterations doesn't change, we compute the last ``belief" that transmitted to $r$ as $r's$ score map. This procedure can be efficiently implemented by distance transform \cite{dt}.

\textit{iii)} After we get the score maps from \textit{ii)}, we further compute the score maps for nodes in the upper layers of ST-AOG. By this procedure, the score map of the root Or-node for neighbouring frames $I_i$ and $I_{i+1}$ can be computed by maximizing Eqn. (\ref{eqn:score_o}) on each spatial-temporal point.

\textit{iv)} For each spatial-temporal sliding window with score being greater than the threshold $\tau$, we follow the Breadth-First-Search (BFS) order to retrieve its parse tree (including the whole car windows, semantic part windows and part statuses at each frame) by taking $\argmax$ at each encountered Or-node. 

\vspace{-2mm}
\subsection{Post-Refinement}	\label{sec:hmm}
\vspace{-1mm}
As analysed in \cite{nbest}, we can get more accurate part localizations by generating multiple object instances in an image. In this paper, we utilize a similar method to generate multiple car part proposals: First of all, we set low detection threshold to generate multiple car proposals around the target car, then we execute a conservative non-maximum suppression (NMS) procedure to override the observation term and select windows that are not local maxima. By backtracing the parse tree $pt$, we can get the semantic part proposals from each car proposal.

The right of Fig. \ref{fig:lbp_hmm} illustrates our part-based HMM inference method. For a semantic part $p$, we can write its bounding box sequence in a given video ($N$ frames) as a stochastic set $\Omega_p = \{p^i\}^{N-1}_{i=1}$, then the overall linking scores of part $p$ in a video can be computed as:
\vspace{-3mm}
\small
\begin{align}
& S(\Omega_p) = \sum_{i=1}^{N-1} S(p^i) + \theta^p_{\mathcal{M}} \psi(p^{i,i+1})
+ \lambda ov(p^i, p^{i+1}) \label{eqn:part_obj}
\end{align}
\normalsize
\vspace{-4mm}
\footnotesize
\begin{align}
& \nonumber	S(p^i) = \theta_p \phi_p(I_i) + \theta_p \phi_p(I_{i+1}), \psi(p^{i,i+1})=(dx,dx^2,dy,dy^2,ds,ds^2)
\vspace{-3mm}
\end{align}
\normalsize
where we model $S(p^i)$ as the appearance and deformation score, $\theta_p$ and $\phi_p$ are the corresponding parameter and feature, $\theta^p_{\mathcal{M}}$ is the temporal weights, $\psi$ is the temporal feature, $dx,dy,ds$ are the differences of temporal position and scale of part $p$ between $I_i$ and $I_{i+1}$, $ov(p^i, p^{i+1})$ is the bounding box overlap of $p$ between $I_i$ and $I_{i+1}$.
For each semantic part in the video, we seek the optimal path $\Omega_p^*$ by maximizing Eqn. (\ref{eqn:part_obj}). 

\vspace{-2mm}
\section{Learning by Latent Structural SVM}
\vspace{-1mm}
\textit{The training data.}  In our car fluent dataset, for each video $X_v, v \in \{1, \cdots, W\}$, we have its annotated parse graph $pg(X_v)$ including the car bounding boxes, viewpoints, car types, semantic part bounding boxes, part statuses and corresponding fluent labels. 

For parameter optimization, the objective function of ST-AOG is:
\vspace{-3mm}
\small
\begin{align}
&\min_{\Theta,\xi} \quad \mathcal{E}(\Theta) = \frac{1}{2} \Arrowvert \Theta \Arrowvert^2 + C \sum_{v=1}^W \xi_v
\vspace{-1mm}
\label{eqn:obj}
\end{align}
\normalsize

\vspace{-6mm}
\small
\begin{align}
\nonumber & \mbox{s.t.} \quad  \xi_v > 0; \qquad \forall v, \quad \widehat{pg}(X_v) \neq pg(X_v),\\
\nonumber & S(pg(X_v)) - S(\widehat{pg}(X_v)) \geq L(pg(X_v), \widehat{pg}(X_v)) - \xi_v
\vspace{-1mm}
\end{align}
\normalsize
where $C$ is the regularization parameter, $\widehat{pg}$ is the predicted parse graph, and $L(pg(X_v), \widehat{pg}(X_v))$ is the surrogate loss function. To accurately predict the viewpoint, car type, part bounding boxes and part statuses, we specify $L(pg(X_v), \widehat{pg}(X_v)) = {1\over {v_N}}\sum_{j=1}^{v_N} \ell (pt(X^j_v), \widehat{pt}(X_v^j))$ ($v_N$ is the number of frames in $X_v$), and define $\ell(pt, \widehat{pt})$ by,
\vspace{-1.5mm}
\begin{equation}
\footnotesize 
\ell(pt, \widehat{pt}) = \left\{ 
  \begin{array}{l l}
    1 & \quad \text{if view or car type is different}\\
    1 & \quad \text{if $overlap(pt,\widehat{pt}) < ov$}\\
    1 & \quad \text{if $\exists p$, $pt.status(p) \neq \widehat{pt}.satus(p)$}\\   
    0 & \quad \text{otherwise} 
  \end{array} \right., 
\label{eqn:los1}
\vspace{-1mm}
\end{equation} 
where the second term computes the overlap between two parse trees (they have the similar structure, otherwise the loss is computed based on the first term), and $overlap(pt, \widehat{pt})$ is defined by the minimum overlap of bounding boxes of all nodes in the parse trees. The third term check the semantic part status. 

The training procedure has the following two steps:

\textit{i) Initialization}. We use standard SVM to train the semantic car part templates on the annotations. For each view, we first cluster the semantic parts by their status, then cluster the ``open" parts by their aspect ratios, as they have large appearance and geometry variations. After getting these clusters, we train a template for each of them. These templates construct the Terminal-nodes in ST-AOG.

\textit{ii) Learning the whole ST-AOG} under the LSSVM. As the groundtruth annotations may not be the most discriminative, we allow parts to move around but with a certain overlap with the groundtruth. Thus the optimization becomes a latent structural problem. We iterate the following two steps to get a local optima:

\textbf{Step 1 - Positives Relabelling}: Do loss-augmented inference for each training video with current model by:
\vspace{-3mm}
\small
\begin{align}
\Omega_p^* = \argmax {\sum_{p=1}^{P} S(\Omega_p) + L(pg(X_v), \widehat{pg}(X_v))}
\vspace{-1mm}
\end{align}
\normalsize

\textbf{Step 2 - Update Weights}: With the inferred part bounding box and part status sequences, we use standard Structural SVM to optimize the appearance, deformation, and motion parameters $\Theta$ in ST-AOG.

In our implementation, we start with a small number of negative examples, and iteratively increase them in the $2^{nd}$ step as the iCCCP method \cite{leozhu_cvpr10}. Besides, we also maintain a cache for hard negative mining. Since computing on every frame is not necessary, we sliding the spatial-temporal window by every $3$ frames.

\begin{figure*}
\centering
\includegraphics[width = 0.9\textwidth]{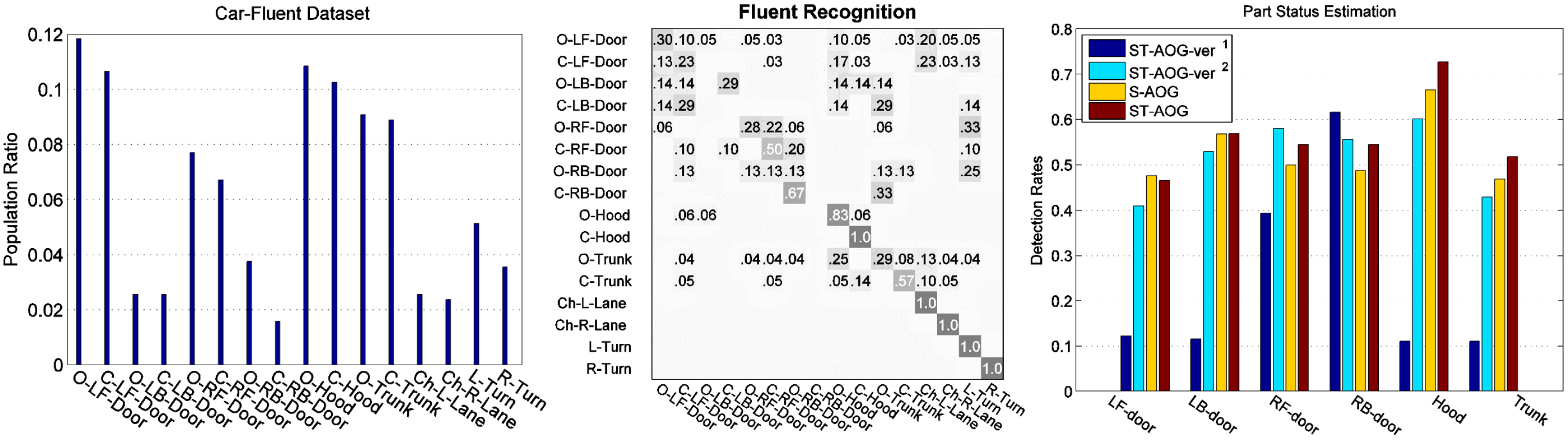}
\caption{Left: semantic fluents and their population ratios on Car-Fluent Dataset. Middle: confusion matrix of the best ST-AOG on this dataset. Right: part status estimation results of different variants of ST-AOG. Here, ``O", ``C", ``Ch", ``L", ``R" denotes ``open", ``close", ``change", ``left", ``right", respectively. \vspace{-4mm} } 
\label{fig:merge} 
\vspace{-1mm}
\end{figure*}

\begin{figure*}
\centering
\includegraphics[width = 0.95\textwidth]{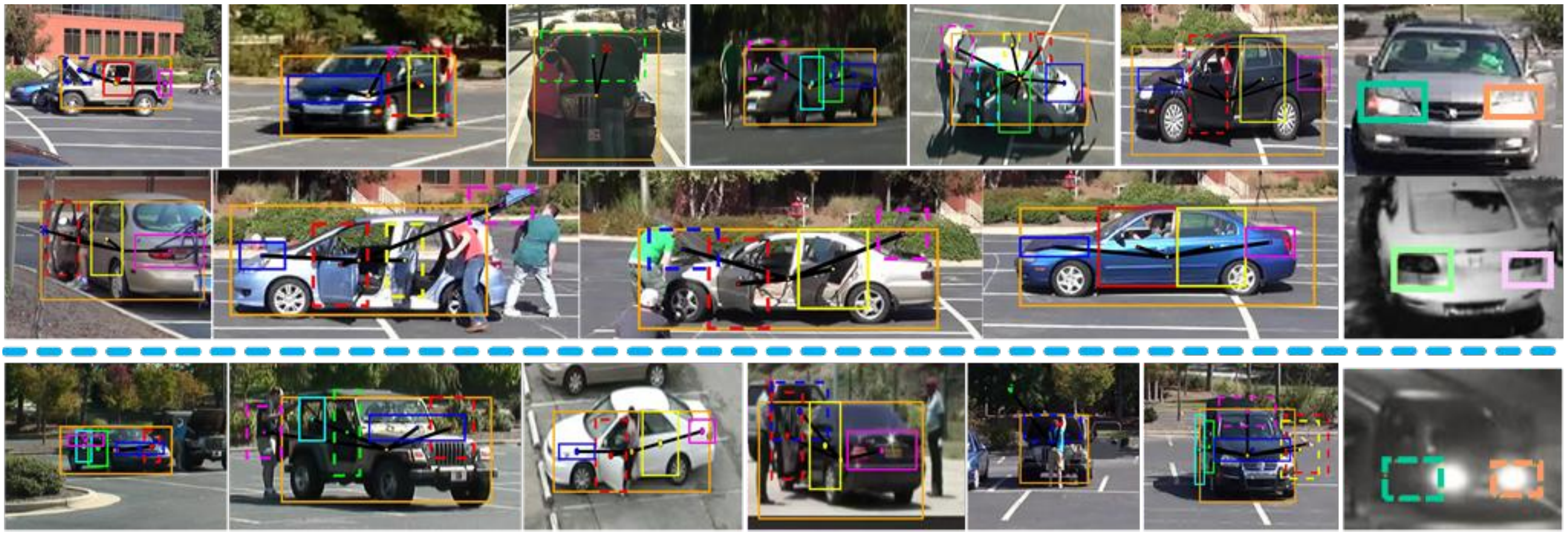}
\caption{Examples of part status estimation results on our Car-Fluent dataset.
For good visualization, we cropped the cars from original images.
Different semantic parts are shown by rectangles in different colors (solid for ``close/off" status and dashed for ``open/on" status).
The black lines are used to illustrate variations of the relative positions between whole car and car parts with different statuses.
The top $2$ rows show successful examples.
The bottom row shows failure cases, which are mainly caused by occlusion and mis-detected car view.
by failure, we mean some part statuses are wrong (e.g., for the 3rd example, the door is correct,
but the hood and trunk are wrong).
Best viewed in color and magnification. \vspace{-4mm} }
\label{fig:dets} 
\end{figure*}

\begin{table*} \scriptsize
\begin{center}
\resizebox{0.80\hsize}{!}{
\begin{tabular}{|c|c|c|c|c|c|c|c|c|c|c|c|}
\hline
{} & \multicolumn{7}{|c|}{BaseLines} & \multicolumn{3}{|c|}{ST-AOG} \\
\hline
{} & STIP-BoW & iDT-BoW & STIP-FV & STIP-VLAD & iDT-FV & iDT-VLAD & C3D & TPS & TPS+iDT & TPS+iDT+C3D\\
\hline
MP & $28.8$ & $35.1$ & $35.2$ & $31.5$ & $40.7$ & $41.3$ & $31.0$ & $34.4$ & $45.0$ & $\mathbf{50.8}$ \\
\hline
\end{tabular}
}
\end{center}
\caption{Results of baseline models and our ST-AOG in terms of mean precision (MP) in \textbf{Car Fluent Recognition}. \vspace{-5mm} }
\label{tab:fluent} 
\vspace{-6mm}
\end{table*}

\vspace{-2mm}
\section{Experiments}
\vspace{-1mm}

\subsection{Car-Fluent Dataset}\label{sec:dataset}
\vspace{-1mm}
The introduced Car-Fluent dataset includes $647$ video clips, containing basically $16$ semantic car part fluents with multiple car types, camera viewpoints and occlusion conditions. All the fluents and its population ratios are shown on the left of Fig. \ref{fig:merge}. The videos in Car-Fluent dataset are collected from various sources (youtube, movies, VIRAT \cite{virat}, etc.), and captured by both static cameras and moving cameras. Based on the number of semantic fluent instances in each video, we split these videos into $2$ sets: the first set consists of $507$ videos where each video contains a single instance of each car fluent. We randomly select $280$ videos in this dataset for training, and the rest for testing; The second split consists of $140$ videos that are just used for testing, in which each video is much longer and has multiple instances of diverse kinds of car fluents. All the videos are used for three important tasks: 1) semantic fluent recognition (e.g. opening hood, change  left lane), 2) car part localization (e.g. where is the hood? Where is the light?), and 3) car part status estimation (e.g., is the trunk open?).

All videos are annotated with viewpoints, car types, bounding boxes of both whole car and semantic parts, and part statuses. The type of part status are {\it ``open/on", ``close/off", and ``occluded"}. In this paper, we focus on ``semantic" and ``functional" parts, all these parts are related to the fluents listed in Fig. \ref{fig:merge}.
We adopt VATIC \cite{vatic} to annotate videos on the introduced dataset. $7$ workers are hired to clip the sequence of semantic car fluents and annotate the videos. Three independent annotations were collected for each video and combined via outlier rejection and averaging. It took them about $500$ hours in total. Finally, all labelling results were thoroughly checked by the authors.
For more details, please refer to our supplementary.

\vspace{-1mm}
\subsection{Fluent Recognition} \label{sec:recog}
\vspace{-1mm}
To verify our ST-AOG on fluent recognition, we compare it with space-time interest points (STIP) \cite{laptev08}, improved dense trajectory (iDT) \cite{idt}, and the deep learning based C3D method \cite{c3d}. For STIP and iDT, we use the softwares in \cite{laptev08,idt} to extract corresponding features inside the car bounding boxes from each video, and try various methods to construct the video descriptors. For the bag-of-words (BoW) method, we try different size of codebooks (i.e., $400, 800, 1000, 1200, 1600, 2000$), and select the best ones for iDT and STIP. For fisher vectors (FV) and  VLAD vectors, we use $256$ components for  as common choices in \cite{event_xu,sanchez13ijcv,JDSP10}.  For feature dimension reduction, we use PCA, and experiment with different reduced dimensions, i.e., $1/4, 1/3, 1/2, 1$ of the original dimension, and utilize the reduced dimensions that best balance the performance and storage cost in corresponding features, i.e., half of the original dimension for both STIP and iDT. 

For C3D \cite{c3d}, since we don't have enough videos to train a deep network, we first pre-train a network on the Sport1M dataset \cite{sport1m} and VIRAT dataset \cite{virat}. Then we finetuned a model on the Car-Fluent dataset. 
As in \cite{c3d}, we extract fc6 features from each video and use the mean features as the video descriptors. We also tried computing the FV and VLAD vectors for C3D method, but get low performance.

For our ST-AOG, based on the predicted semantic part positions and part statuses extracted from parse graph $pg$, we compute temporal part status (TPS) features as follows:
\vspace{-2mm}
\small
\begin{align}
& \nonumber \mathcal{\varphi}_T = \{[\phi_1(p^i_s),\phi_2(p^i_s,p^{i+1}_s),d(p^i,p^{i+1})]_{p=1}^{P}\}_{i=1}^{N-1}
\vspace{-1mm}
\end{align}
\normalsize
where $\phi_1(s) = [0,\cdots,1(s),\cdots,0]$ is a $z_p$ dimension vector, here $z_p$ is the number of statuses of part $p$, $\phi_2(s^i,s^{i+1})=[0,\cdots,1(s^i)1(s^{i+1}),\cdots,0]$ is a $z_p \times z_p$ dimension vector, $d(p^i,p^{i+1})$ is the temporal transition of part $p$ between neighbouring frames.
For each car light, we also compute the mean intensity difference over the light bounding box across neighbouring frames, and append it to $\mathcal{\varphi}_T$.
We compute FV and VLAD vectors on TPS to construct the video descriptors as STIP and iDT. For our TPS, we get the best performance with $53$ reduced feature dimensions for PCA and $256$ centers for VLAD vectors.

We use the one-against-rest approach to train multi-class SVM classifiers for above methods. For evaluation, we plot confusion matrix, and report the mean precision (MP) in fluent recognition, equivalent to the average over the diagonal of the confusion matrix. 

The first $6$ columns in Table \ref{tab:fluent} show the MP of above methods. As can be seen, the ``iDT-VLAD" get the best performance, while C3D and our TPS get relatively lower performances on this dataset, which verified the importance of motion features. For C3D, we think the relatively low performance is due to the lack of training data and it doesn't take advantage of car positions.
As TPS, iDT and C3D belong to different type of features, we augment our TPS by integrating it with iDT and C3D features. To this end, we first combine our TPS with iDT, specifically, we try various combinations of FV and VLAD vectors, and find the best one is concatenating TPS VLAD vectors and iDT VLAD vectors with square root normalization, as can be seen in Table \ref{tab:fluent}, this improves the performance by about $4$; then we further combine TPS and iDT VLAD vectors with C3D feature, as can be seen in the last column, we improve the performance to $50.8$, which outperforms the best single feature by $9.5$. 
From these results we can see, ST-AOG  provides important cues in car fluent recognition, and it's very general to integrate with traditional motion features and deep learning features.

The middle of Fig.~\ref{fig:merge} shows the confusion matrix of our ``TPS+iDT+C3D" method. We can see the merged feature performs pretty well on fluents, i.e., ``Open/Close Hood", ``Change Left/Right Lane", ``Left/Right Turn", but totally miss fluents, e.g., ``Open/Close LB Door", ``Close RB Door". 
For fluents related to the car back door, the low performance may be attributed to their ambiguous features and relatively low population ratios on current dataset.

\begin{table*} \scriptsize
\begin{center}
\resizebox{0.72\hsize}{!}{
\begin{tabular}{|c|c|c|c|c|c|c|}
\hline
{} & SSDPM \cite{ssdpm} & AOG-Car \cite{boli_iccv13} & DP-DPM \cite{dpdpm} & ST-AOG-HOG & ST-AOG-HOG-CSC & ST-AOG-CNN\\
\hline
Body & $96.4$ & $\mathbf{99.1}$ & $93.4$ & $99.0$ & $90.8$ & $94.7$ \\
\hline
Left-front door & $36.3$ & $36.6$ & $14.4$ & $\mathbf{49.9}$ & $48.0$ & $36.6$ \\
\hline
Left-back door & $42.2$ & $44.4$ & $7.1$ & $\mathbf{60.3}$ & $55.1$ & $32.3$ \\
\hline
Right-front door & $31.6$ & $40.7$ & $13.0$ & $\mathbf{58.4}$ & $55.9$ & $26.8$ \\
\hline
Right-back door & $33.0$ & $19.4$ & $14.2$ & $\mathbf{61.4}$ & $55.3$ & $32.8$ \\
\hline
Hood & $35.9$ & $20.1$ & $7.1$ & $\mathbf{73.7}$ & $67.5$ & $38.4$ \\
\hline
Trunk & $25.4$ & $16.8$ & $10.1$ & $\mathbf{54.0}$ & $49.7$ & $33.7$ \\
\hline
Left-head Light & $10.4$ & $22.1$ & $19.6$ & $\mathbf{33.1}$ & $27.8$ & $29.3$ \\
\hline
Right-head Light & $13.3$ & $24.3$ & $15.7$ & $\mathbf{41.3}$ & $36.5$ & $23.3$ \\
\hline
Left-tail Light & $6.8$ & $25.7$ & $18.3$ & $\mathbf{27.8}$ & $27.3$ & $22.7$ \\
\hline
Right-tail Light & $6.2$ & $\mathbf{31.1}$ & $14.0$ & $23.2$ & $22.9$ & $14.0$ \\
\hline
Mean & $24.1$ & $28.1$ & $13.4$ & $\mathbf{48.3}$ & $44.6$ & $29.0$ \\
\hline
\end{tabular}
}
\end{center}
\caption{Semantic \textbf{Car Part Localization} results of baseline methods and ST-AOG on Car-Fluent dataset. \label{tab:part_loc}}
\vspace{-5mm}
\end{table*}

\subsection{Part Localization and Part Status Estimation}
\vspace{-1mm}
Since part location and status are important to fluent recognition, in this experiment, we further examine the frame-level part localization and status estimation results.
\vspace{-8mm}
\subsubsection{Part Localization}
\vspace{-1mm}
For semantic part localization, we use following baselines: 
\textbf{(i)} \textit{Strongly supervised deformable part-based model (SSDPM) \cite{ssdpm}}. SSDPM is an open source software, it  achieved state-of-the-art performance on localizing semantic parts of animals on PASCAL VOC 2007 dataset \cite{pascal}.
\textbf{(ii)} \textit{And-Or Structures \cite{boli_iccv13}}. In the literature, \cite{boli_iccv13} also attempted to localize the semantic parts by replacing the latent ones in DPM. 
\textbf{(iii)} \textit{Deep DPM (DP-DPM) \cite{dpdpm}}. The authors in \cite{dpdpm} argued that by replacing HOG features with Deep CNN features can improve the detection performance of original DPM \cite{DPM}. We use the same trick as \cite{boli_iccv13} to add semantic parts on DP-DPM.
We also try to compare our model with \cite{ramananPose,HR_car,Chen_CVPR14,Chen_NIPS14}, but find it's hard to compare with them, as either the code is not available, or the design of parts is inappropriate in our case (details are in the  supplementary).

For our ST-AOG, we use several variants for comparison. 1)``ST-AOG-HOG" is the instance of our ST-AOG based on HOG features. 2) ``ST-AOG-HOG-CSC" is the cascade version of our model. To improve the speed, we use the recently developed Faster-RCNN \cite{faster_rcnn} for cascade detection of cars. Then based on these car detections, we further run the ST-AOG for part localization. 3) ``ST-AOG-CNN" is the instance of our ST-AOG based on the pyramid of CNN features, here, we use the $max_5$ layer of the CNN architecture for comparison as \cite{dpdpm}. To cope with small cars, we double the size of source images for CNN.

For evaluation, we compute the detection rate of the semantic parts (e.g., ``left-front door", ``hood").
Here, we assume the whole car bounding boxes are given, and each method only outputs $1$ prediction with the highest confidence for each groundtruth car. 
Table \ref{tab:part_loc} shows the quantitative results. We can see on almost all semantic parts, our model outperforms baseline models by a large margin. We believe this is because ST-AOG jointly model the view, part occlusion, geometry and appearance variation, and temporal shifts. For ``ST-AOG-HOG-CSC", we just use it as a reference to improve the speed of ST-AOG, since it has the nearly real-time speed, and can be plugged in our model as an outlier-rejection module.
Surprisingly, CNN features perform worse than HOG features on the Car-Fluents dataset. In experiments, we find the extracted CNN feature on deep pyramid \cite{dpdpm} is too coarse (the cell size is 16), even we resize original images by $2$ times, it still miss many small parts. Based on recent study of using CNN for keypoints prediction \cite{malik_CVPR15}, we believe a more effective method is required to integrate CNN feature with graphical models.

\vspace{-4mm}
\subsubsection{Part Status Estimation}
\vspace{-1mm}
To the best of our knowledge, no model tried to output car part status in our case in the literature. To evaluate the part status, we compare different versions of our model and analyse the effects of different components. Specifically, we compute the part status detection rates, that is, a part status is correct if and only if its position and status (e.g., {\it open, close}) are both correct. 

On the right of Fig.~\ref{fig:merge}, we show the quantitative results of $4$ versions of our ST-AOG. Here, we use HOG features for simplicity. ``ST-AOG-ver$^1$" refers to the ST-AOG without both view and part status penalty in Eqn. \eqref{eqn:los1}, ``ST-AOG-ver$^2$" refers to the ST-AOG without view penalty in Eqn. \eqref{eqn:los1}, and ``ST-AOG" is the full model with all penalty in Eqn. \eqref{eqn:los1}. To investigate the effect of motion flow, we also compare the ST-AOG without motion flow, i.e., the ``S-AOG". As can be seen, the view penalty is the most important factor, then is the part status and motion flow, and part status seems to have more importance. Interestingly, for some parts (e.g., ``RB-Door"), adding more penalty or temporal info will decrease the performance.

Fig.~\ref{fig:dets} shows some qualitative running examples of ST-AOG. The first two rows show the successful results, the last row shows the failure examples.
For better visualization, we cropped the image to just contain the interested car.
We can see our model can localize parts and recognize their statuses fairly good with different viewpoints and car types, but may report wrong results when people occlude the car, the viewpoints are mis-detected, or there are background clutters.


\vspace{-2mm}
\section{Conclusion}
\vspace{-1mm}
This paper proposed a ST-AOG to recognize car fluents at semantic part-level from video. The ST-AOG integrates the motion features in temporal, and a LBP-based DP method can be used in inference.  The model parameters are learned under the latent structural SVM (LSSVM) framework. 
To promote the research of fluents, we have collected a new Car-Fluent dataset with detailed annotations of viewpoints, car types, part bounding boxes, and part statuses. This dataset presents new challenges to computer vision, and complements existing benchmarks. Our experiments verified the ability of proposed ST-AOG on fluent recognition, part localization, and part status estimation. 

In future work, we will integrate human and car jointly, and study human-car interactions to help understand human actions/intents based on car fluent recognition.


\noindent\textbf{Acknowledgement:} B. Li is supported by China 973 Program under Grant NO. 2012CB316300. T.F. Wu and S.C. Zhu are supported by DARPA SIMPLEX Award N66001-15-C-4035, ONR MURI project N00014-16-1-2007, and NSF IIS 1423305. We thank Tianming Shu, Xiaohan Nie and Dan Xie for helpful discussions.

{\small
\bibliographystyle{ieee}
\bibliography{aog_fluent}
}

\newpage
\appendix
\section{Main Challenges on Car-Fluent Dataset} \label{sec:dataset}
\begin{figure*}
\centering
\includegraphics[width = 0.96\textwidth]{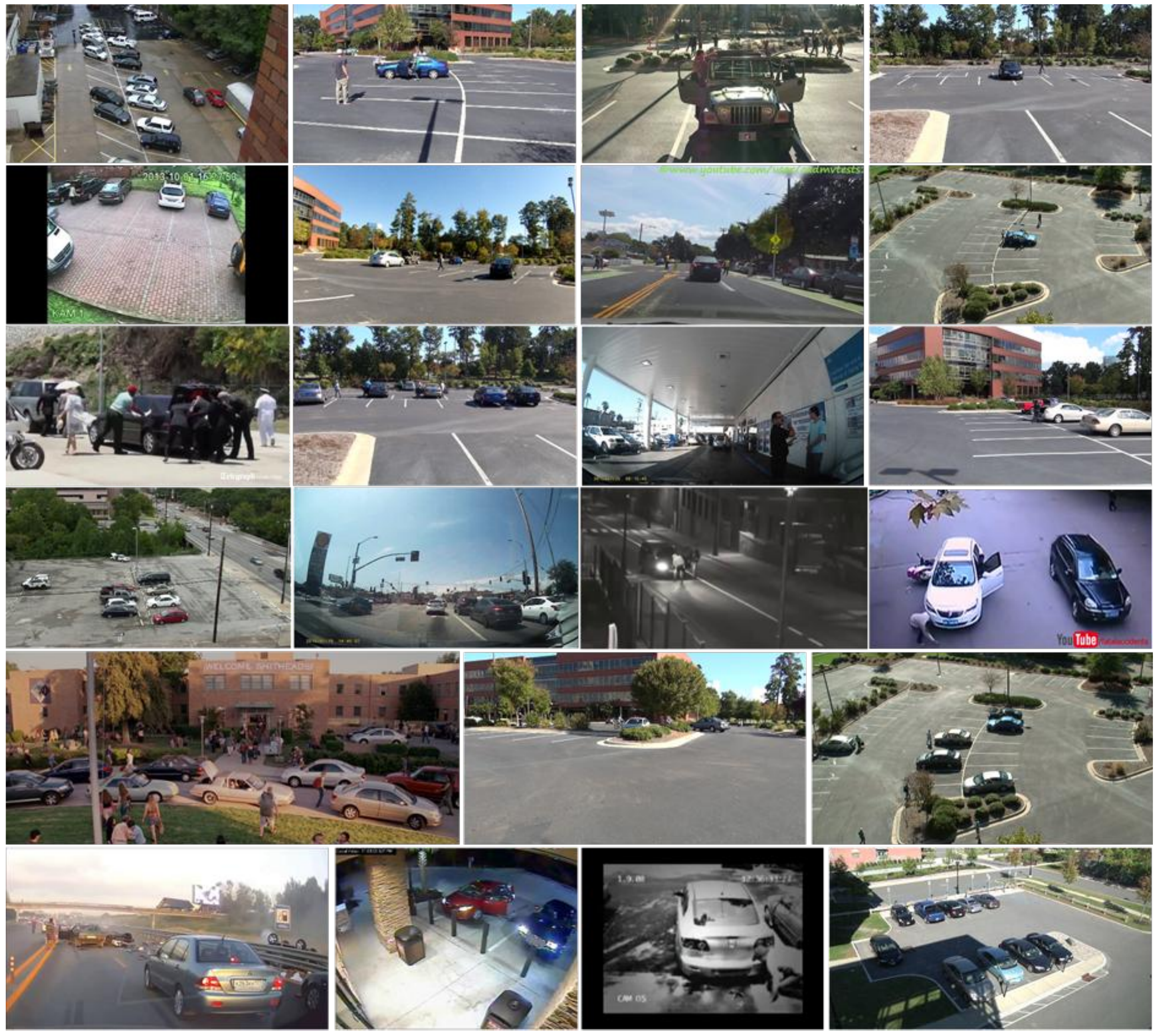}
\caption{Sample images of our Car-Fluent Dataset, from which we can see the diverse scene context of car fluents. }
\label{fig:supp_snap} 
\end{figure*}

Our Car-Fluent dataset includes $647$ video clips, containing basically $10$ types of semantic parts and $16$ types of car part fluents with diverse camera viewpoints and occlusion conditions. Fig. \ref{fig:supp_snap} shows the whole scene context of these videos. 
The videos are collected from various sources (youtube, movies, VIRAT \cite{virat}, etc.), and captured by both static cameras and moving cameras.
As can be seen from Fig. \ref{fig:supp_snap}, there are both high and low resolution parts, which pose great challenges on part localization and status estimation.

On this dataset, the \textbf{semantic parts} are: 
\begin{itemize}
\item ``hood"
\item ``left-front door"
\item ``left-back door"
\item ``right-front door"
\item ``right-back door"
\item ``trunk"
\item ``left head light"
\item ``right head light"
\item ``left tail light"
\item ``right-tail light".
\end{itemize} 

The \textbf{car fluents} are: 
\begin{itemize}
\item ``open/close left-front door" 
\item ``open/close left-back door"
\item ``open/close right-front door" 
\item ``open/close right-back door"
\item ``open/close hood"
\item ``open/close trunk"
\item ``change left/right lane"
\item ``turn left/right".
\end{itemize}

Main challenges on this dataset are all related to car parts and fluents, including\footnote{For car lights, main challenges are the low resolution and ambiguous appearance, as they don't have the geometry change, but have periodically intensity change, we omit these parts in the following analysis.}:
\begin{enumerate}
\item the large geometry and appearance variation of cars introduced by part status change; 
\item low resolution of car parts; 
\item diverse occlusion introduced by people;
\item the variation of fluent execution rate;
\item diverse viewpoints.
\end{enumerate}

\begin{figure*}
\centering
\includegraphics[width = 1.0\textwidth]{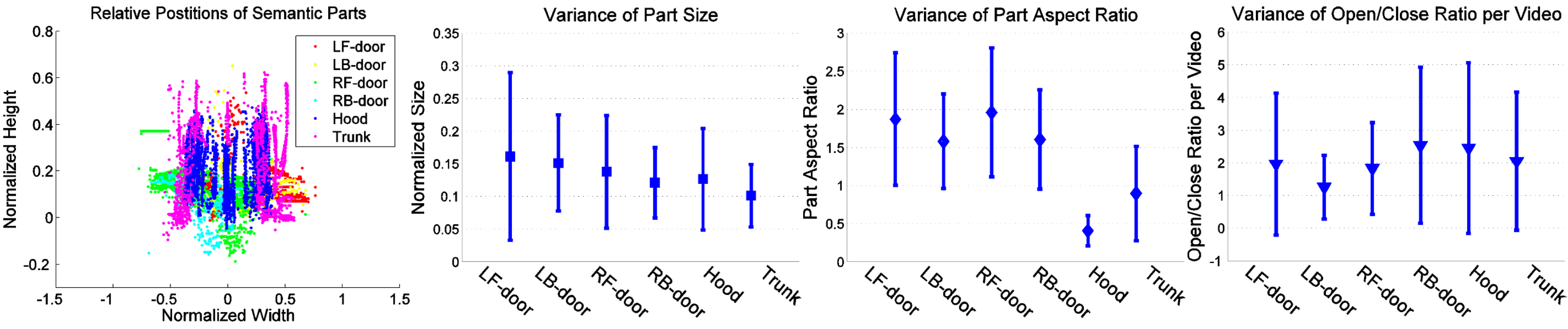}
\caption{Some statistics of semantic parts on Car-Fluent dataset. In the first figure, different parts are coded with different colors. 
In the second and third figures, we show the variances of part size and aspect ratios. These two statistics reflect the geometry variations of car parts on Car-Fluent dataset.
The last figure shows the ``open/close" ratios of car parts, which reflects the diverse execution rates of car fluents.
}
\label{fig:dataset_stats} 
\end{figure*}

The first three figures in Fig.~\ref{fig:dataset_stats} show the relative positions of different semantic parts (which are color coded), the variance of part size, and the variance of part aspect ratio on this dataset. Here, part position and part size are normalized by the size of the whole car, i.e., car body. As can be seen, each semantic part has disordered distribution and large variances of part size, or part aspect ratio. This is because an opened part is very different from a closed one in terms of size, aspect ratio, and relative positions w.r.t car body. 

\begin{figure*}
\centering
\includegraphics[width = 0.96\textwidth]{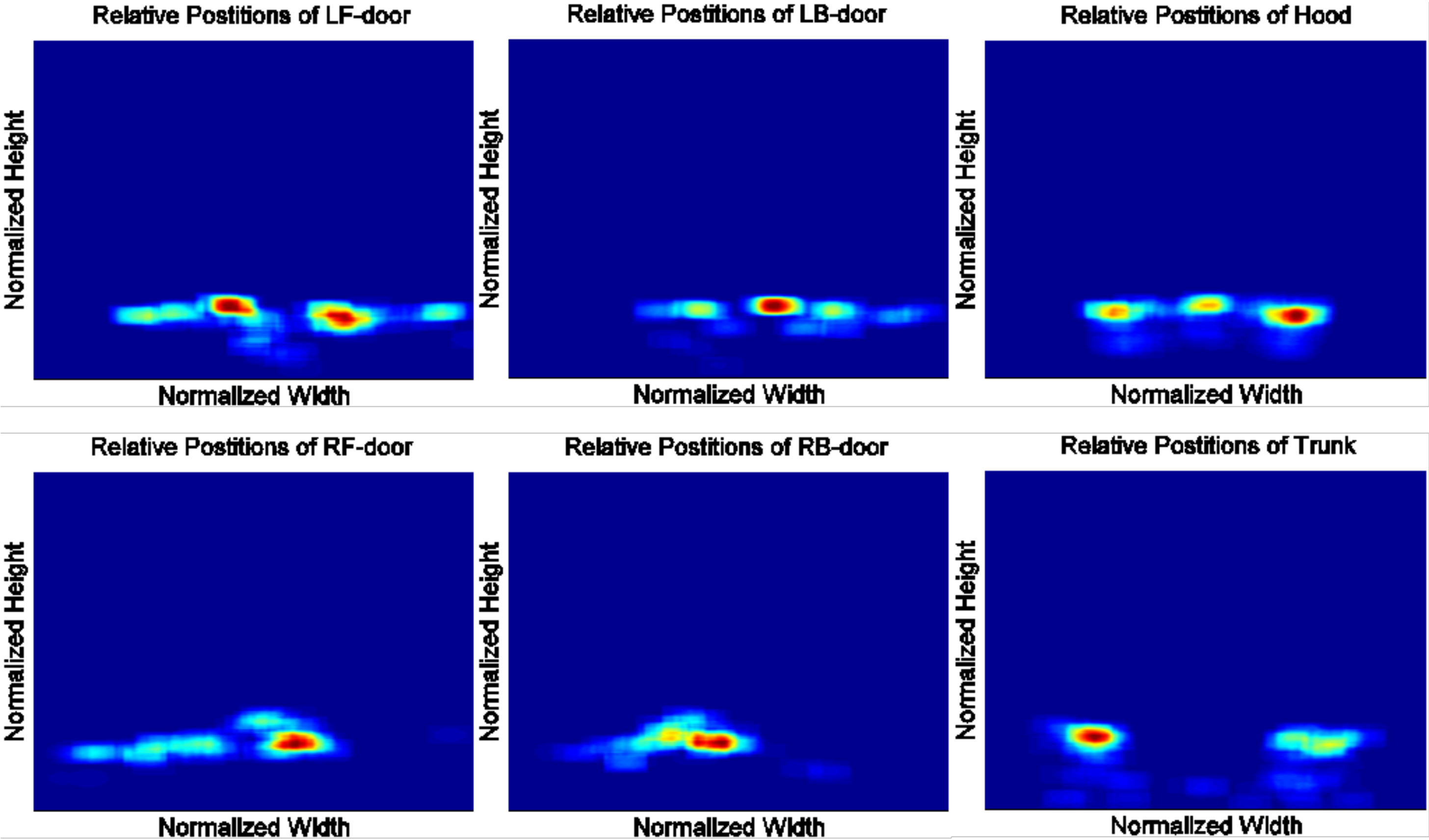}
\caption{Semantic part distributions on Car-Fluent dataset. We align each semantic part with the whole car   position, and normalize the part size with the whole car bounding box.
We can see several ``peaks" in the distribution of each semantic part. These ``peaks" reflect the principal positions of ``open" parts and ``close" parts.
}
\label{fig:heatmap} 
\end{figure*}

For each semantic car part, we also plot the heat map of its distribution, which can be seen in Fig. \ref{fig:heatmap}. We can see the parts are distributed diversely because of the status change, and there are several ``peaks" reflect the principal positions of the ``open" parts and ``close" parts.

On the last of Fig.~\ref{fig:dataset_stats}, we show the number of opening frames compared to the ones of closing frames, which can be viewed as the temporal variance of fluent videos on proposed dataset. The big variance is caused by the diverse execution rate. For instance, the speed of opening left-front door depends on different people, and even for the same person, the speed is not always equal.
Other challenges include intra-class variations of fluent change on different car types (e.g., opening the trunk of a jeep is very different from the same fluent on a sedan) and background clutters.

Based on the experimental results in our  paper, we can see these challenges pose a hard problem to current vision models. Since these videos are captured from real scenarios, we believe it is suitable for fluent recognition and part status estimation in the wild, and hope this could draw more attentions in our community.

\section{Car-Fluent Videos}
We show the challenging videos on Car-Fluent dataset, please check them in the directory - ``video-demos".
Since there are many cars that don't have fluent change in the original video, we ask the annotators to only annotate cars that have fluent change. 
To simplify the annotation, annotators may choose not to annotate the moving cars in the parking lot scenario.

For each video, we show the ``close/turn-off" parts by solid rectangle, the ``open/turn-on" parts by dashed rectangle, and the process of car fluents change by dotted rectangle. When there are fluents change, we also show the text of specific fluents name on the right of each video.

For the car lights video, we only show the whole car bounding boxes for better visualization.

\section{Current Performance}
As can be also seen Table 1 in our paper, the overall performance of fluent recognition is still low. In our experiments, we find our model detect some wrong part status, or just missed when parts are heavily occluded or too small, and thus get wrong transition spatial-temporal features. For STIP and iDT, we find they missed capturing part motions on some videos, and many features are often extracted on people. This probably because car parts are small, ambiguous with background, and often occluded by people. Overall, more informative features and modelling strategies are needed to cope with these cases in the future.

\section{About Part Localization Baselines}
We also try to compare our model with Yang and Ramanan's mixtures-of-parts model \cite{ramananPose}, and its CNN extension \cite{Chen_NIPS14}, but find it's hard to compare with them. There are mainly $3$ reasons:
first, the parts in their model are equally-sized. However, in real life, different semantic car parts have different sizes and aspect ratios, especially when the parts are opened. We found it's not easy to extend their framework to model diverse sizes and aspect ratios of parts; second, it's not very easy to model a fully-connected skeleton model as human pose for car. For example, when facing a car in the frontal view, it is very hard to speculate the bounding box of the tail lights. Thus it's hard to design a fully-connected skeleton model for car. Third, they need fine scale landmark annotations for input while there is no such annotation on Car-Fluents Dataset, besides, parts in their model are not ``functional" (e.g, may not be open).

In literature,  \cite{Chen_CVPR14} also proposed a fully-collected part model which  is related to our work.  But their code is not released, and we found it's not easy to re-implement their  model.

\section{More Part-Localization and Part-Status Estimation Results}
In addition to the results of part localization and status estimation in our paper, we show more results here.

\begin{figure*}
\centering
\includegraphics[width = 1.0\textwidth]{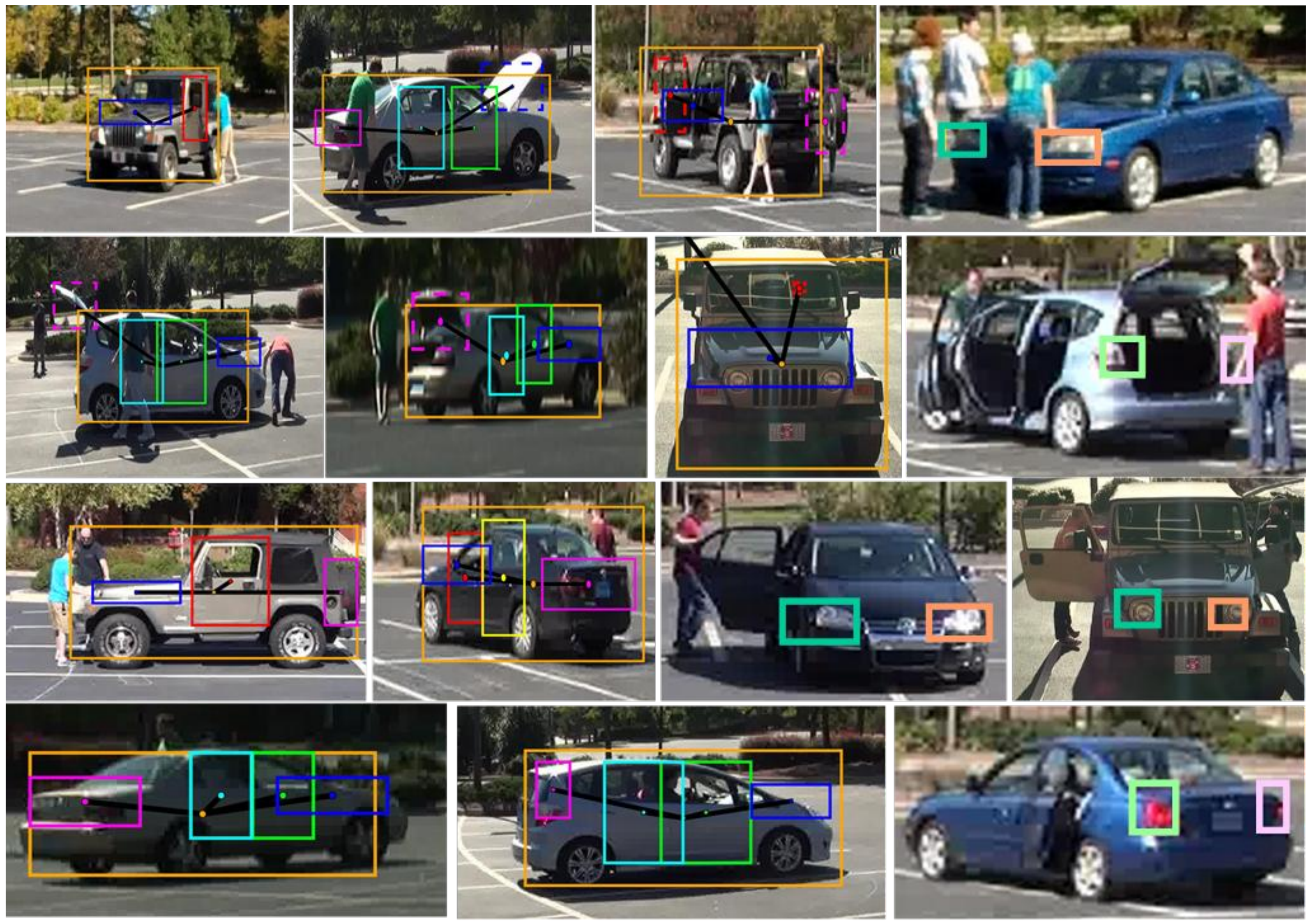}
\caption{Successful detection examples of ST-AOG on Car-Fluent dataset. 
For good visualization, we crop the cars from the original detection images, and show the car lights separately.
Different semantic parts are shown by rectangles in different colors (solid for ``close", or ``turn-off" status, and dashed for ``open", or ``turn-on" status).
As can be seen our ST-AOG is fair in localizing these parts and estimating the corresponding statuses.
\vspace{-3mm} }
\label{fig:supp_succ} 
\end{figure*}

\begin{figure*}
\centering
\includegraphics[width = 1.0\textwidth]{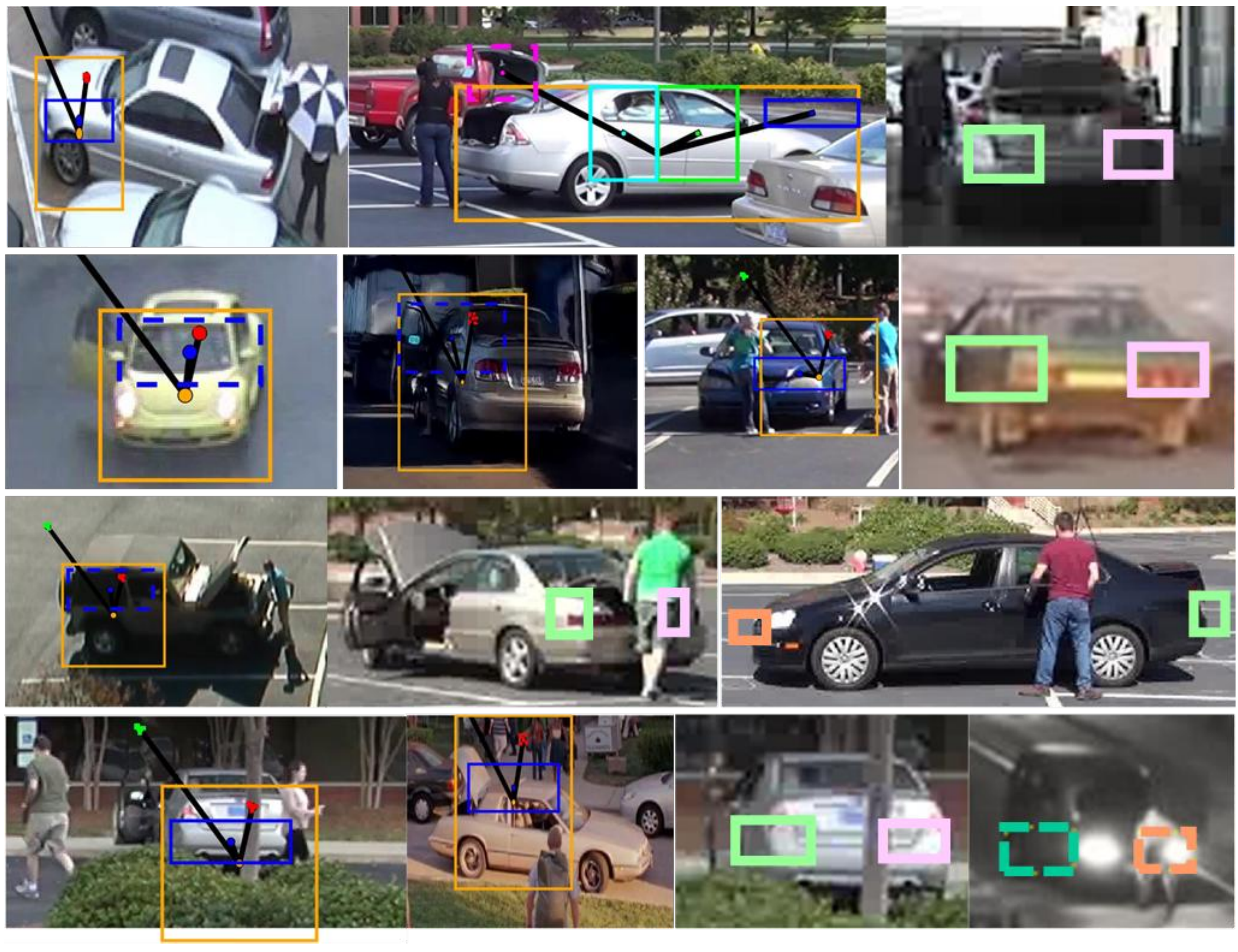}
\caption{Failure detection examples of ST-AOG. 
For good visualization, we crop the cars from the original detection images, and show the car lights separately.
Different semantic parts are shown by rectangles in different colors (solid for ``close", or ``turn-off" status, and dashed for ``open", or ``turn-on" status).
The black lines are used to illustrate variations of the relative positions between whole car and car parts with different statuses.
The failure cases are mainly due to the occlusions introduced by people, the mis-detected viewpoints, low part resolutions, or there are background clutters.
\vspace{-3mm} }
\label{fig:supp_fail} 
\end{figure*}

Fig. \ref{fig:supp_succ} shows more successful examples of our ST-AOG. Fig. \ref{fig:supp_fail} shows more failure ones.
For better visualization, we only show the cropped detection results of cars.
We can see our model can localize parts and estimate their statuses fairly well with different viewpoints and car types.
The failure cases are mainly due to occlusion, car view mis-dectection, low part resolution, or background clutters. 
We will cope with these problems in the future work.
From these detection results, we can also see the position of a part can change a lot when its status change.

\newpage

\end{document}